\documentclass[11pt,a4paper]{article}
\usepackage[hyperref]{acl2019}

\usepackage{times}
\usepackage{latexsym}

\usepackage{amsmath, amsthm, amssymb, amsfonts}
\usepackage{algorithm}
\usepackage{algpseudocode}

\usepackage{helvet}  
\usepackage{courier}  
\usepackage{url}  
\usepackage{graphicx}  
\usepackage{color}

\usepackage{multirow}
\usepackage{booktabs}
\usepackage{comment}
\usepackage{caption}
\usepackage{subcaption}

\aclfinalcopy 
\def\aclpaperid{1972} 

\title{Unsupervised Pivot Translation for Distant Languages}

\author{Yichong Leng\thanks{\ \ \ The work was done when the first author was an intern at Microsoft Research Asia.} \\
  University of Science and Technology of China \\
  \texttt{lyc123go@mail.ustc.edu.cn} \\\And
  Xu Tan \\
  Microsoft Research \\
  \texttt{xuta@microsoft.com} \\\AND
  Tao Qin\thanks{\ \ \ Corresponding author} \\
  Microsoft Research \\
  \texttt{taoqin@microsoft.com} \\\And
  Xiang-Yang Li\footnotemark[2] \\
  University of Science and Technology of China \\
  \texttt{xiangyangli@ustc.edu.cn} \\\AND
  Tie-Yan Liu \\
  Microsoft Research \\
  \texttt{tyliu@microsoft.com}
  }

\date{}

\begin{document}
\maketitle

\begin{abstract}
Unsupervised neural machine translation (NMT) has attracted a lot of attention recently. While state-of-the-art methods for unsupervised translation usually perform well between similar languages (e.g., English-German translation), they perform poorly between distant languages, because unsupervised alignment does not work well for distant languages. In this work, we introduce unsupervised pivot translation for distant languages, which translates a language to a distant language through multiple hops, and the unsupervised translation on each hop is relatively easier than the original direct translation. We propose a learning to route (LTR) method to choose the translation path between the source and target languages. LTR is trained on language pairs whose best translation path is available and is applied to the unseen language pairs for path selection. Experiments on 20 languages and 294 distant language pairs demonstrate the advantages of the unsupervised pivot translation for distant languages, as well as the effectiveness of the proposed LTR for path selection. Specifically, in the best case, LTR achieves an improvement of 5.58 BLEU points over the conventional direct unsupervised method.

\end{abstract}

\section{Introduction}
\label{intro}
Unsupervised neural machine translation (NMT)~\citep{artetxe2017unsupervised,lample2017unsupervised,DBLP:conf/emnlp/LampleOCDR18}, which uses only monolingual sentences for translation, is of great importance for zero-resource language pairs.  Unsupervised translation relies on unsupervised cross-lingual word alignment or sentence alignment ~\citep{conneau2017word,artetxe2017learning}, where word embedding mapping~\citep{artetxe2017unsupervised,lample2017unsupervised} and vocabulary sharing~\citep{DBLP:conf/emnlp/LampleOCDR18} are used for word alignment, and encoder/decoder weight sharing~\citep{artetxe2017unsupervised, DBLP:conf/emnlp/LampleOCDR18} and adversarial training~\citep{lample2017unsupervised} are used for sentence alignment.

Unsupervised cross-lingual alignment works reasonably well for a pair of similar languages, such as English-German or Portuguese-Galician, since they have similar lexica and syntax and share the same alphabets and language branch. However, the alignment between a pair of distant languages, which are not in the same language branch\footnote{In this work, we use language branch to determine distant languages. We choose the taxonomy of language family provided by Ethnologue~\citep{paul2009ethnologue} (https://www.ethnologue.com/browse/families), which is one of the most authoritative and commonly accepted taxonomies. Distant languages can also be defined using other principles, which we leave to future work.}, such as Danish-Galician is challenging. As a consequence, unsupervised translation between distant languages is usually of lower quality. For example, the unsupervised NMT model achieves 23.43 BLEU score on Portuguese-Galician translation, while just 6.56 on Danish-Galician translation according to our experiments. In this work, we focus on unsupervised translation of distant languages. 

We observe that two distant languages can be linked through multiple intermediate hops where unsupervised translation of two languages on each hop is easier than direct translation of the two distance languages, considering that the two languages on each intermediate hop are more similar, or the size of monolingual training data is larger. Therefore, we propose unsupervised pivot translation through multiple hops for distant languages, where each hop consists of unsupervised translation of a relatively easier language pair. For example, the distant language pair Danish$\to$Galician can be translated by three easier hops: Danish$\to$English, English$\to$Spanish and Spanish$\to$Galician. In this way, unsupervised pivot translation results in better accuracy (12.14 BLEU score) than direct unsupervised translation (6.56 BLEU score) from Danish to Galician in our experiments. 

The challenge of unsupervised pivot translation is how to choose a good translation path. Given a distant language pair X$\to$Y, there exists a large amount of paths that can translate from X to Y\footnote{Suppose we only consider translation paths with at most $N$ hops. Given $M$ candidate intermediate languages, there are $\frac{M!}{(M-N+1)!}$ possible paths.}, and different paths may yield very different translation accuracies. Therefore, unsupervised pivot translation may result in lower accuracy than direct unsupervised translation if a poor path is chosen. How to choose a path with good translation accuracy is important to guarantee the performance of unsupervised pivot translation. 

A straightforward method is to calculate the translation accuracies of all possible paths on a validation set and choose the path with the best accuracy. However, it is computationally unaffordable due to the large amount of possible paths. For example, suppose we consider at most 3 hops ($N=3$) and 100 languages ($M=100$), and assume each path takes an average of 20 minutes to translate all the sentences in the validation set using one NVIDIA P100 GPU to get the BLEU score. Then it will take nearly 1400000 GPU days to evaluate all candidate paths. Even if we just consider 20 languages ($M=20$), it will still take 2200 GPU days. Therefore, an efficient method for path selection is needed. We propose a learning to route (LTR) method that adopts a path accuracy predictor (a multi-layer LSTM) to select a good path for a distant language pair. Given a translation path and the translation accuracy of each hop on the path, the path accuracy predictor can predict the overall translation accuracy following this path.  Such a predictor is first trained on a training set of paths with known overall accuracy, and then used to predict the accuracy of a path unseen before.

We conduct experiments on a large dataset with 20 languages and a total of 294 distant language pairs to verify the effectiveness of our method. Our proposed LTR achieves an improvement of more than 5 BLEU points on some language pairs. 

The contributions of this paper are as follows: (1) We introduce pivot translation into unsupervised NMT to improve the accuracy of distant languages. (2) We propose the learning to route (LTR) method to automatically select the good translation path. (3) Large scale experiments on more than 20 languages and 294 distant language pairs demonstrate the effectiveness of our method.

\section{Related Work}
In this section, we review the related work from three aspects: unsupervised neural machine translation, pivot translation, and path routing.

\paragraph{Unsupervised NMT} As the foundation of unsupervised sentence translation, unsupervised word alignment has been investigated by~\citep{conneau2017word,artetxe2017learning}, where linear embedding mapping and adversarial training are used to ensure the distribution-level matching, achieving considerable good accuracy or even surpasses the supervised counterpart for similar languages. \citet{artetxe2017unsupervised,lample2017unsupervised} propose unsupervised NMT that leverages word translation for the initialization of the bilingual word embeddings. ~\citet{DBLP:conf/acl/XuYCW18} partially share the parameter of the encoder and decoder to enhance the semantic alignment between source and target language. ~\citet{DBLP:conf/emnlp/LampleOCDR18} further share the vocabulary of source and target languages and jointly learned the word embeddings to improve the quality of word alignment, and achieve large improvements on similar language pairs. Recently, inspired by the success of BERT~\citep{devlin2018bert} and MASS~\citep{song2019mass}, ~\citet{song2019mass} leverage the MASS pre-training in the unsupervised NMT model and achieve state-of-the-art performance on some popular language pairs.

Previous works on unsupervised NMT can indeed achieve good accuracy on similar language pairs, especially on the closely related languages such as English and German that are in the same language branch. In this circumstance, they can simply share the vocabulary and learn joint BPE for source and target languages, and share the encoder and decoder, which is extremely helpful for word embedding and latent representation alignment. However, they usually achieve poor accuracy on distant languages that are not in the same language branch or do not share same alphabets. In this paper, we propose pivot translation for distant languages, and leverage the basic unsupervised NMT model in~\citep{DBLP:conf/emnlp/LampleOCDR18} on similar languages as the building blocks for the unsupervised pivot translation.

\paragraph{Pivot Translation} Pivot translation has long been studied in statistical machine translation to improve the accuracy of low/zero-resource translation~\citep{wu2007pivot,utiyama2007comparison}.
\citet{cheng2017joint,chen2017teacher} also adapt the pivot based method into neural machine translation. However, conventional pivot translation usually leverages a resource-rich language (mainly English) as the pivot to help the low/zero-resource translation, while our method only relies on the unsupervised method in each hop of the translation path. Due to the large amount of possible path choices, the accuracy drops quickly along the multiple translation hops in the unsupervised setting, unsupervised pivot translation may result in lower accuracy if the path is not carefully chosen. In this situation, path selection (path routing) will be critical to guarantee the performance of pivot translation.

\paragraph{Path Routing}
Routing is the process of selecting a path for traffic in a network, or between or across multiple networks. Generally speaking, routing can be  performed in many types of networks, including circuit switching network~\citep{girard1990routing}, computer networks (e.g., Internet)~\citep{huitema2000routing}, transportation networks~\citep{raff1983routing} and social networks~\citep{liben2005geographic}. In this paper, we consider the routing of the translation among multiple languages, where the translation accuracy is the criterion for the path selection. Usually, the translation accuracy of the multi-hop path is not simply the linear combination of the accuracy on each one-hop path, which makes it difficult to route for a good path.

\section{Unsupervised Pivot Translation}
\label{sec_multi_pivot}
Observing that unsupervised translation is usually hard for distant languages, we split the direct translation into multiple hops, where the unsupervised translations on each hop is relatively easier than the original direct unsupervised translation. Formally, for the translation from language $X$ to $Y$, we denote the pivot translation as
\begin{equation}
 X\to Z_1 \to ... \to Z_n \to Y,
\end{equation}
where $Z_1$, ..., $Z_n$ are the pivot languages and $n$ is the number of pivot languages in the path. We set $n \in \{0, 1, 2\}$ and consider 3-hop path at most in this paper, considering the computation cost and accuracy drop in longer translation path. Note that when $n=0$, it is the one-hop (direct) translation.

There exists a large amount of translation paths between $X$ and $Y$ and each path can result in different translation accuracies, or even lower than the direct unsupervised translation, due to the information loss along the multiple translation hops especially when unsupervised translation quality on one hop is low. Therefore, how to choose a good translation path is critical to ensure the accuracy of unsupervised pivot translation. In this section, we introduce the learning to route (LTR) method for the translation path selection.

\subsection{Learning to Route}  
\label{LTR}
In this section, we first give the description of the problem formulation, and then introduce the training data, features and model used for LTR.

\paragraph{Problem Formulation} We formulate the path selection as a translation accuracy prediction problem. The LTR model learns to predict the translation accuracy of each path from language $X$ to $Y$ given the translation accuracy of each hops in the path, and the path with the highest predicted translation accuracy among all the possible paths is chosen as the output. 

\paragraph{Training Data} We construct the training data for the LTR model in the following steps:
(1) From $M$ languages, we choose the distant language pairs whose source and target languages are not in the same language branch. We then choose a small part of the distant language pairs as the development/test set respectively for LTR, and regard the remaining part as the training set for LTR. (2) In order to get the accuracy of different translation paths for the distant language pairs, as well as to obtain the input features for LTR, we train the unsupervised model for the translation between any languages and obtain the BLEU score of each pair. For $M$ languages, there are total $M(M-1)$ language pairs and BLEU scores, which requires $M(M-1)/2$ unsupervised models since one model can handle both translation directions following the common practice in unsupervised NMT~\citep{DBLP:conf/emnlp/LampleOCDR18}.
(3) We then test the BLEU scores of the each possible translation path for the language pairs in the development and test sets, based on the models trained in previous steps. These BLEU scores are regarded as the ground-truth data to evaluate the performance of unsupervised pivot translation. (4) We just get the BLEU scores of a small part of the possible paths in the training set, which are used as the training data for LTR model\footnote{As described in Footnote 2, we cannot afford to test the BLEU scores of all the possible paths, so we just test a small part of them for training.}. We describe the features of the training data in the next paragraph.

\paragraph{Features}  We extract several features from the paths in training data for better model training. Without loss of generality, we take a 3-hop path $X\to Z_{1}\to Z_{2}\to Y$ as an example, and regard it as a token sequence consisting of languages and one-hops: $X$, $X\to Z_{1}$, $Z_{1}$, $Z_{1}\to Z_{2}$, $Z_{2}$, $Z_{2}\to Y$ and $Y$. We consider two kinds of features for the token sequence: (1) The token ID. There are a total of 7 tokens in the path shown above. Each token ID is converted into trainable embeddings. For a one-hop token like $Z_{1}\to Z_{2}$, its embedding is simply the average of the two embeddings of $Z_{1}$ and $Z_{2}$. (2) The BLEU score of each language and one-hop, where we get the BLEU score of each language by averaging the accuracy of the one-hop path from or to this language. For example, the BLEU score of the target language $Z_{1}$ in $X\to Z_{1}$ is calculated by averaging all the BLEU scores of the one-hop translation from other languages to $Z_{1}$, while the BLEU score of the source language $Z_{1}$ in $Z_{1}\to Z_{2}$ is calculated by averaging the BLEU scores of all the one-hop translation from $Z_{1}$ to other languages. We concatenate the above two features together in one vector for each language and one-hop token, and get a sequence of features for each path. The BLEU score of the path will be used as the label for the LTR model.

\paragraph{Model}  We use a multi-layer LSTM model to predict the BLEU score of the translation path. The input of the LSTM model is the feature sequence described in the above paragraph. The last hidden of LSTM is multiplied with an one-dimensional vector to predict the BLEU score of the path.

\subsection{Discussions}
\label{discussion}
We make brief discussions on some possible baseline routing methods and compare them with our proposed LTR.

\textit{Random Routing}: We randomly choose a path as the routing result.

\textit{Prior Pivoting}: We set the pivot language for each language according to prior knowledge\footnote{For the languages in each language branch, we choose the language with the largest amount of monolingual data in this branch as the pivot language. All languages in the same language branch share the same pivot language.}. Denote $P_{X}$ and $P_{Y}$ as the pivot language for $X$ and $Y$ respectively. The path $X\to P_{X}\to P_{Y}\to Y$ will be chosen as the routing result by prior pivoting.

\textit{Hop Average}: The average of the BLEU scores of each one-hop in the path is taken as the predicted BLEU score for this path. We select the path with the highest predicted BLEU score, as used in the LTR method.  

Compared with these simple rule based routing methods described above, LTR chooses the path purely by learning on a part of the ground-truth paths. The feature we designed in LTR can capture the relationship between languages to determine the BLEU score and relative ranking of the paths. This data-driven learning based method (LTR) will be more accurate than the rule based methods. In the next section, we conduct experiments to verify effectiveness of our proposed LTR and compare with the baseline methods.

\section{Experiments Design}
Our experiments consist of two stages in general. In the first stage, we need to train the unsupervised NMT model between any two languages to get the BLEU scores of each one-hop path. We also get the BLEU scores for a part of multi-hop paths through pivoting, which are used as the training and evaluation data for the second stage. In the second stage, we train the LTR model based on the training data generated in the first stage. In this section, we give brief descriptions of the experiment settings for the unsupervised NMT model training (the first stage) and the LTR model training and path routing (the second stage).

\subsection{Experiment Setting for Direct Unsupervised NMT}
\label{sec_exp_setting_unsup}
\paragraph{Datasets}
We conduct the experiments on 20 languages and a total of 20$\times$19$=$380 language pairs, which have no bilingual sentence pairs but just monolingual sentences for each language.
The languages involved in the experiments can be divided into 4 language branches by the taxonomy of language family: Balto-Slavic branch, Germanic branch, Italic branch and Uralic branch\footnote{The first three branches belong to Indo-European family while the last branch is actually a language family. We do not further split the 3 languages in Uralic family into different branches.}. The language name and its ISO 639-1 code contained in each branch can be found in the supplementary materials (Section 1 and 2).

We collect the monolingual corpus from Wikipedia for each language.  We
download the language specific Wikipedia contents in XML format\footnote{For example, we download English Wikipedia contents from https://dumps.wikimedia.org/enwiki.}, and use WikiExtractor\footnote{https://github.com/attardi/wikiextractor} to
extract and clean the texts. We then use the sentence tokenizer from the NLTK toolkit\footnote{https://www.nltk.org/} to generate segmented sentences from Wikipedia documents.

To ensure we have the development and test set for the large amount of language pairs to evaluate the unsupervised translation accuracy in our experiments, we choose the languages that are covered by the common corpus of TED talks, which contains translations between more than 50 languages~\citep{Ye2018WordEmbeddings}\footnote{https://github.com/neulab/word-embeddings-for-nmt}. In this circumstance, we can leverage the development and test set from TED talks for evaluation. Note that in the unsupervised setting, we just leverage monolingual sentences for unsupervised training and only use the bilingual data for developing and testing. In order to alleviate the domain mismatch problem that we train on monolingual data from Wikipedia but test on the evaluation data from TED talks, we also fine-tune the unsupervised models with the small size of monolingual data from TED talks\footnote{https://github.com/ajinkyakulkarni14/TED-Multilingual-Parallel-Corpus/tree/master/Monolingual\_data}. The monolingual data from TED talks is merged with the  monolingual data from Wikipedia in the fine-tuning process, which results in better performance for the unsupervised translation. The size of Wikipidia and TED talks monolingual data can be found in the supplementary materials (Section 3).

All the sentences in the bilingual and monolingual data are first tokenized with moses tokenizer\footnote{https://github.com/moses-smt/mosesdecoder/blob/mast er/scripts/tokenizer/tokenizer.perl} and then segmented into subword symbols using Byte Pair Encoding (BPE)~\citep{DBLP:conf/acl/SennrichHB16a}. When training the unsupervised model, we learn the BPE tokens with 60000 merge operations across the source and target languages for each language pair and jointly training the embedding using fastext\footnote{https://github.com/facebookresearch/fastText}, following the practice in~\citet{DBLP:conf/emnlp/LampleOCDR18}.

\paragraph{Model Configurations}
We use transformer model as the basic NMT model structure, considering it achieves state-of-the-art accuracy and becomes a popular choice for recent NMT research. We use 4-layer encoder and 4-layer decoder with model hidden size $d_{\text{model}}$ and feed-forward hidden size $d_{\text{ff}}$ being 512, 2048 following the default configurations in~\citet{DBLP:conf/emnlp/LampleOCDR18}. We use the same model configurations for all the language pairs.

\paragraph{Model Training and Inference}
We train the unsupervised model with 1 NVIDIA Tesla V100 GPU. One mini-batch contains roughly 4096 source tokens and 4096 target tokens, as used in ~\citet{DBLP:conf/emnlp/LampleOCDR18}. We follow the default parameters of Adam optimizer~\citep{kingma2014adam} and learning rate schedule in~\citet{DBLP:conf/nips/VaswaniSPUJGKP17}. During inference, we decode with greedy search for all the languages. We evaluate the translation quality by tokenized case sensitive BLEU~\citep{DBLP:conf/acl/PapineniRWZ02} with multi-bleu.pl\footnote{https://github.com/moses-smt/mosesdecoder/blob/ master/scripts/generic/multi-bleu.perl}.

\begin{table*}[t]
\small
\centering
\begin{tabular}{c  c c | c c c c | c c c c }
\toprule
Source & Target & \textit{DT} & \textit{GT} & \textit{GT}($\Delta$) & Pivot-1 & Pivot-2 & \textit{LTR} & \textit{LTR}($\Delta$) & Pivot-1 & Pivot-2 \\
\midrule
Da & Gl & 6.56 & 12.14 & 5.58 & En & Es & 12.14 & 5.58 & En & Es \\
Bg & Sv & 4.72 &  9.92 & 5.20 & En & En &  9.92 & 5.20 & En & En \\
Gl & Sv & 3.79 &  8.62 & 4.83 & Es & En &  8.62 & 4.83 & Es & En \\
Sv & Gl & 3.70 &  8.13 & 4.43 & En & Es &  8.13 & 4.43 & En & Es \\
Be & It & 2.11 &  6.40 & 4.29 & Uk & En &  5.24 & 3.13 & En & En \\
Pt & Be & 4.76 &  8.86 & 4.10 & Ru & Ru &  8.86 & 4.10 & Ru & Ru \\
Gl & Da & 7.45 & 11.33 & 3.88 & Es & Es & 11.33 & 3.88 & Es & Es \\
Be & Pt & 6.39 &  9.77 & 3.38 & Ru & Ru &  6.39 & 0.00 & -   & -   \\
It & Be & 2.24 &  5.19 & 2.95 & Pt & Ru &  4.64 & 2.40 & Ru & Ru \\
Nl & Uk & 4.69 &  7.23 & 2.54 & De & De &  7.12 & 2.53 & Ru & Ru \\

\bottomrule
\end{tabular}
\caption{The BLEU scores of a part of the distant language pairs in the test set (Please refer to Section 1 and 4 in the supplementary materials for the corresponding full language name and full results). \textit{DT}: direct unsupervised translation. \textit{GT}: the ground-truth best path. \textit{LTR}: the routing results of LTR. ($\Delta$) is the BLEU gap between \textit{GT} or \textit{LTR} and \textit{DT}. Pivot-1 and Pivot-2 are two pivot languages in the path, which will be the same language if the path is 2-hop and will both be empty if the path is 1-hop (direct translation).} 
\label{test_set_bleu_score_number}
\end{table*}

\begin{table}[h]
\small
\centering
\begin{tabular}{c c c c }
\toprule
Length & \textit{1-hop} & \textit{2-hop } & \textit{3-hop } \\
\midrule
Ratio (\%)  &  7.1 & 53.6 & 39.3 \\
\bottomrule
\end{tabular}
\caption{The length distribution of the best translation paths. The ratio is calculated based on all language pairs in the test set.} 
\label{unsup_best_setting_distribution}
\end{table}

\subsection{Experiment Setting for Routing} 
\paragraph{Configurations for Routing}
We choose the distant language pairs from the 20 languages based on the taxonomy of language family: if two languages are not in the same language branch, then they are regarded as distant languages. We get 294 distant language pairs. As described in Section~\ref{LTR}, we choose nearly 5\% and 10\% of the distant language pairs as the development and test set for routing. Note that if the language pair $X\to Y$ is in development (test) set, then the language pair $Y\to X$ will be also in development (test) set. We then enumerate all possible paths between any two language pairs in development and test set, and test the BLEU scores of the each possible path, which are regarded as the ground-truth data to evaluate the performance of the routing method. For the remaining 85\% distant language pairs, we just test the BLEU score for 10\% of all possible paths, and use these BLEU scores as the label for LTR model training. 

We use 2-layer LSTM as described in Section~\ref{LTR}. The dimension of input feature vector is 6, which includes the embedding of the token ID with size of 5, the BLEU score with size 1 (we normalize the BLEU score into 0-1). We change the depth and width of LSTM, but there is no significant gain in performance.

We use the mean square error as the training loss for the LTR model, and use Adam as the optimizer. The initial learning rate is 0.01. When applying the LTR model on unseen pairs, we predict the BLEU scores of all the possible paths (including 1-hop (direct translation), 2-hop and 3-hop translation path) between the source and target languages, and choose the path with the highest predicted BLEU score as the routing result. Note that when predicting the path with LTR in inference time, we do not include the pivot language which occurs less than 10 times in training set, which can improve that stability of the \textit{LTR} prediction.

\paragraph{Methods for Comparison}
We conduct experimental comparisons on different methods described in Section~\ref{sec_multi_pivot} for path selection (routing), including \textit{Random Routing (RR)}, \textit{Prior Pivoting (PP)}, \textit{Hop Average (HA)} and \textit{Learning to Route (LTR)}. We also compare these routing methods with the direct unsupervised translation, denoted as \textit{Direct Translation (DT)}. We list the BLEU score of the best multi-hop path (the ground truth) as a reference, which is denoted as \textit{Ground Truth (GT)}.

\section{Results}
In this section, we introduce the performance of unsupervised pivot translation for distant languages. We first demonstrate the advantages of unsupervised pivot translation by comparing the best translation path (\textit{GT}) with direction translation (\textit{DT}), and then show the results of our proposed \textit{LTR}. We also compare \textit{LTR} with other routing methods (\textit{RR}, \textit{PP} and \textit{HA}) to demonstrate its effectiveness.

\subsection{The Advantage of Unsupervised Pivot Translation}
In order to demonstrate the advantage of unsupervised pivot translation for distant languages, we first analyze the distribution of the length of the best translation paths (\textit{GT}), as shown in Table~\ref{unsup_best_setting_distribution}. The direction translation (1-hop) only takes a ratio of 7.1\%, which means that a majority (92.9\%) of the distant language pairs need multiple hops to improve the translation accuracy.

\begin{figure}[t]
\small
\centering
    \includegraphics[width=0.5\textwidth]{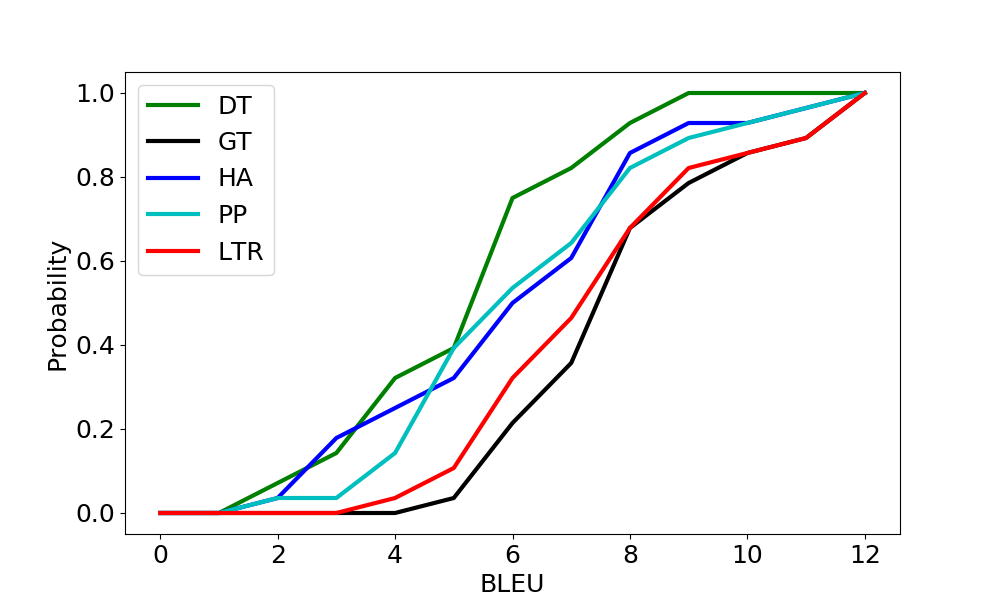}
    \caption{The CDF of the BLEU scores for the distant language pairs in the test set. The green curve represents the direct unsupervised translation (\textit{DT}), and the black curve represents the best translation path (\textit{GT}). The other three curves represent the three routing methods for comparison: blue for hop average (\textit{HA}), cyan for prior pivoting (\textit{PP}) and red for our proposed learning to route (\textit{LTR}). } 
\label{fig_cdf}
\end{figure}

We further compare the BLEU score of the best path (\textit{GT}, which is also the upper-bound of different routing methods) with the direct unsupervised translation, and show the results for a part of distant languages pairs in Table~\ref{test_set_bleu_score_number}\footnote{Due to space limitation, we leave the full results of the distant language pairs in the test set in the supplementary materials (Section 4).}. It can be seen that \textit{GT} can largely outperform the direct translation \textit{DT} with up to 5.58 BLEU points. We further plot the CDF of the BLEU scores on all the distant language pairs in the test set in Figure~\ref{fig_cdf}. It can be seen that the CDF curve of \textit{GT} is always in the right part of \textit{DT}, which means better accuracy and demonstrates the advantage of unsupervised pivot translation for distant languages.

\subsection{Results of \textit{LTR}}

\paragraph{Accuracy of \textit{LTR} Model} As our \textit{LTR} selects the good path by ranking according to the predicted BLEU score, we first report the accuracy of selecting the best path. \textit{LTR} can achieve 57\% in terms of top-1 accuracy and 86\% in terms of top-5 accuracy. Although the top-1 accuracy is not so high, it is acceptable because there exists some other route path just a little bit lower than the best path. We show the routing results of our \textit{LTR} for some language pairs in Table~\ref{test_set_bleu_score_number}. Take the \textit{Nl-Uk} language pair in Table~\ref{test_set_bleu_score_number} as an example. The routing result of \textit{LTR} for this pair does not match with \textit{GT}, which affects the top-1 accuracy. However, the BLEU gap between our selected path and the best path is as small as 0.09, which has little influence on the BLEU score of the selected path. Our further analysis in the next paragraph shows that the averaged BLEU score that \textit{LTR} achieved in test set is close to that of \textit{GT}.

\begin{table}[t]
\small
\centering
\begin{tabular}{c| c| c c c c| c }
\toprule
Methods & \textit{DT} & \textit{RR} & \textit{HA} & \textit{PP} & \textit{LTR} & \textit{GT}  \\
\midrule
BLEU & 6.06 & 3.40 & 6.92 & 7.12 & 8.33 & 8.70 \\
\bottomrule
\end{tabular}
\caption{The performance of different routing methods. The BLEU score is averaged on all the distant language pairs in the test set. The compared methods include: \textit{DT}: direct unsupervised translation, \textit{RR}: random routing, \textit{PP}: prior pivoting, \textit{HA}: hop average, \textit{LTR}: our proposed learning to route, and \textit{GT}: the best translation path (the ground truth).} 
\label{unsup_methods_performance}
\end{table}

\paragraph{BLEU Score of Selected Path}
We further report the BLEU score of the translation path selected by \textit{LTR} as well as other routing methods in Table~\ref{unsup_methods_performance}, where the BLEU score is averaged over all the distant language pairs in the test set. It can be seen that compared with direct unsupervised translation (\textit{DT}) which achieves 6.06 averaged BLEU score\footnote{The averaged BLEU score seems not high, because the unsupervised translations between some hard languages in the test set obtain really low BLEU scores, which affects the average score.}, our \textit{LTR} can achieve 2.27 BLEU points improvement on average, and is just 0.37 points lower than the ground truth best path (\textit{GT}). The small gap between the ground truth and \textit{LTR} demonstrates that although \textit{LTR} fails to select the best path in 43\% of the distant pairs (just 57\% in terms of top-1 accuracy), it indeed chooses the path which has a BLEU score slightly lower than the best path. Random routing (\textit{RR}) even performs worse than \textit{DT}, demonstrating the routing problem is non-trivial. \textit{LTR} outperforms \textit{PP} and \textit{HA} by more than 1 BLEU point on average. We also show the CDF of the BLEU scores of different methods in Figure~\ref{fig_cdf}, which clearly shows that \textit{LTR} can outperform the \textit{PP} and \textit{HA} routing methods, demonstrating the effectiveness of the proposed \textit{LTR}.

\subsection{Extension to Supervised Pivoting}

\begin{table*}[t]
\small
\centering
\begin{tabular}{c  c c c c  c  c | c c c c c  c  c}
\toprule
Source & Target & \textit{DT} & \textit{GT}-unsup & \textit{GT}-sup & $\Delta$ && &Source & Target & \textit{DT} & \textit{GT}-unsup & \textit{GT}-sup & $\Delta$ \\
\midrule
Da & Gl & 6.56 & 12.14 & 15.20 & 8.64 &&& Pt & Be & 4.76 &  8.86 & 13.03 & 8.27 \\
Bg & Sv & 4.72 &  9.92 &  9.92 & 5.20 &&& Gl & Da & 7.45 & 11.33 & 15.52 & 8.07 \\
Gl & Sv & 3.79 &  8.62 &  9.58 & 5.79 &&& Be & Pt & 6.39 &  9.77 & 14.50 & 8.11 \\
Sv & Gl & 3.70 &  8.13 &  9.38 & 5.68 &&& It & Be & 2.24 &  5.19 &  8.60 & 6.36 \\
Be & It & 2.11 &  6.40 &  9.26 & 7.15 &&& Nl & Uk & 4.69 &  7.23 &  8.07 & 3.38 \\

\bottomrule
\end{tabular}
\caption{The BLEU scores of the same language pairs as shown in Table \ref{test_set_bleu_score_number} (Please refer to Section 5 in the supplementary materials for the full results of the test set). \textit{GT}-sup and \textit{GT}-unsup represent the ground-truth best path with and without supervised pivoting. $\Delta$ is the BLEU gap between \textit{GT}-sup and \textit{DT}.} 
\label{hub_sup_test_set_bleu_score_number}
\end{table*}

\begin{table}[t]
\small
\centering
\begin{tabular}{c| c| c c c c| c }
\toprule
Methods & \textit{DT} & \textit{RR} & \textit{HA} & \textit{PP} & \textit{LTR} & \textit{GT}  \\
\midrule
BLEU & 6.06 & 3.46 & 7.07 & 8.84 & 9.45 & 9.79 \\
\bottomrule
\end{tabular}
\caption{The performance of different routing methods when enhanced with supervised pivoting. The BLEU score is averaged on all the distant language pairs in the test set. The compared methods include: \textit{DT}: direct unsupervised translation, \textit{RR}: random routing, \textit{HA}: hop average, \textit{PP}: prior pivoting, \textit{LTR}: our proposed learning to route, and \textit{GT}: the best translation path (the ground truth).} 
\label{hubsup_methods_performance}
\end{table}

In the previous experiments, we rely purely on unsupervised NMT for pivot translation, assuming that the translation on each hop cannot leverage any bilingual sentence pairs.  However, there indeed exist plenty of bilingual sentence pairs between some languages, especially among the popular languages of the world, such as the official languages of the United Nations and the European Union. If we can rely on some supervised hop in the translation path, the accuracy of the translation for distant languages would be greatly improved. 

Take the translation from Danish to Galician as an example. The BLEU score of the direct unsupervised translation is 6.56, while the ground-truth best unsupervised path (Danish$\to$ English$\to$Spanish$\to$Galician) can achieve a BLEU score of 12.14, 5.58 points higher than direct unsupervised translation. For the translation on the intermediate hop, i.e, English$\to$Spanish, we have a lot of bilingual data to train a strong supervised translation model. If we replace the unsupervised English$\to$Spanish translation with the supervised counterpart, the BLEU score of the path (Danish$\to$ English$\to$Spanish$\to$Galician) can improve from 12.14 to 15.2, with 8.64 points gain over the direct unsupervised translation. Note that the gain is achieved without leveraging any bilingual sentence pairs between Danish and Galician.

Without loss of generality, we choose 6 popular languages (we select English, German, Spanish, French, Finish and Russian to cover each language branch we considered in this work) as the supervised pivot languages and replace the translations between these languages with the supervised counterparts. Note that we do not leverage any bilingual data related to the source language and target languages, and the supervised models are only used in the intermediate hop of a 3-hop path. For the bilingual sentence pairs between pivot languages, we choose the common corpus of TED talk which contains translations between multiple languages~\citep{Ye2018WordEmbeddings}\footnote{This is the same dataset where we choose the development and test sets in Section~\ref{sec_exp_setting_unsup}. The data can be downloaded from https://github.com/neulab/word-embeddings-for-nmt.}.

Table~\ref{hub_sup_test_set_bleu_score_number} shows the performance improvements on the language pairs (the same pairs as shown in Table \ref{test_set_bleu_score_number}). When enhanced with supervised pivoting, we can achieve more than 8 BLEU points gain over \textit{DT} on 4 language pairs, without using any bilingual data between the source language or target language. We also compare our proposed learning to route method \textit{LTR} with \textit{RR}, \textit{HA} and \textit{PP}, as showed in Table~\ref{hubsup_methods_performance}. We conduct the experiments on the original development and test set, but removing the language pairs whose source and target languages belong to the supervised pivot languages we choose. It can be seen that \textit{LTR} can still outperform \textit{RR}, \textit{HA} and \textit{PP} and be close to \textit{GT}, demonstrating the effectiveness of \textit{LTR} in the supervised pivoting setting.

\section{Conclusions and Future Work}
In this paper, we have introduced unsupervised pivot translation for distant language pairs, and proposed the learning to route (LTR) method to automatically select a good translation path for a distant language pair. Experiments on 20 languages and totally 294 distant language pairs demonstrate that (1) unsupervised pivot translation achieves large improvements over direct unsupervised translation for distant languages; (2) our proposed LTR can select the translation path whose translation accuracy is close to the ground-truth best path; (3) if we leverage supervised translation instead of the unsupervised translation for some popular language pairs in the intermediate hop, we can further boost the performance of unsupervised pivot translation.

For further works, we will leverage more supervised translation hops to improve the performance of unsupervised translation for distant languages. We will extend our method to more distant languages.

\bibliography{acl2019}
\bibliographystyle{acl_natbib}

\end{document}